\documentclass[manuscript,screen]{acmart}
\AtBeginDocument{%
  \providecommand\BibTeX{{%
    \normalfont B\kern-0.5em{\scshape i\kern-0.25em b}\kern-0.8em\TeX}}}

\setcopyright{rightsretained}
\copyrightyear{2026}
\acmYear{2026}
\acmConference[SIGGRAPH Posters '26]{Special Interest Group on Computer Graphics and Interactive Techniques Conference Posters}{July 19--23, 2026}{Los Angeles, CA, USA}
\acmBooktitle{Special Interest Group on Computer Graphics and Interactive Techniques Conference Posters (SIGGRAPH Posters '26), July 19--23, 2026, Los Angeles, CA, USA}
\acmDOI{10.1145/nnnnnnn.nnnnnnn} 
\acmISBN{978-x-xxxx-xxxx-x/26/07} 

\usepackage{colortbl}
\usepackage{xcolor}
\usepackage{enumitem}
\usepackage{float}
\definecolor{lightgreen}{rgb}{0.8,0.93,0.8}
\definecolor{lightgray}{rgb}{0.83,0.83,0.83}

\copyrightyear{2026}
\acmYear{2026}
\acmConference[SIGGRAPH Posters '26]{Special Interest Group on Computer Graphics and Interactive Techniques Conference Posters}{July 19--23, 2026}{Los Angeles, CA, USA}
\acmBooktitle{Special Interest Group on Computer Graphics and Interactive Techniques Conference Posters (SIGGRAPH Posters '26), July 19--23, 2026, Los Angeles, CA, USA}
\acmDOI{10.1145/3799825.3818741}
\acmISBN{979-8-4007-2548-7/2026/07}

\begin{document}

\title{Cage-based Texture Transfer with Geometric Filtering}

\author{Rose Mei Zhou}
\affiliation{%
  \institution{Roblox}
   \country{USA}
}
\email{rosezhou@roblox.com}

\author{Lynnette Hui Xian Ng}
\affiliation{%
  \institution{CMU}
   \country{USA}
}
\email{lynnetteng@cmu.edu}

\author{Adrian Xuan Wei Lim}
\affiliation{%
  \institution{Roblox}
   \country{USA}
}
\email{xlim@roblox.com}

\author{Conor Griffin}
\affiliation{%
  \institution{Roblox}
   \country{USA}
}
\email{cgriffin@roblox.com}

\author{Faraz Baghernezhad}
\affiliation{%
  \institution{Roblox}
   \country{USA}
}
\email{fbaghernezhad@roblox.com}

\renewcommand{\shortauthors}{Zhou et al.}

\begin{abstract}
Real-time texture transfer expands the creative horizon for interactive applications, enabling seamless detail projection in scenarios that range from digital character cosmetics to procedural automotive texturing. Yet, its practical application is governed by inherent trade-offs between processing speed and suppression of artifacts. Low-latency transfer methods frequently fail to suppress artifacts, and robust alternatives rely on large-scale models that are costly in training and memory. Our proposed method bridges the gap between efficiency and robustness by using a cage-based geometric filtering method to identify Non-Cosmetic Zones (NCZs) for artifact suppression. While other models are resource-intensive and require multiple days of training on manually annotated datasets, we are able to successfully suppress artifacts and achieve immediate deployment on consumer-grade hardware. Our framework achieved highly efficient runtimes of $\sim~$70ms on mobile devices for a $\sim4.8k$ triangle mesh. 
\end{abstract}



\begin{teaserfigure}
\centering
    \includegraphics[scale=0.60]{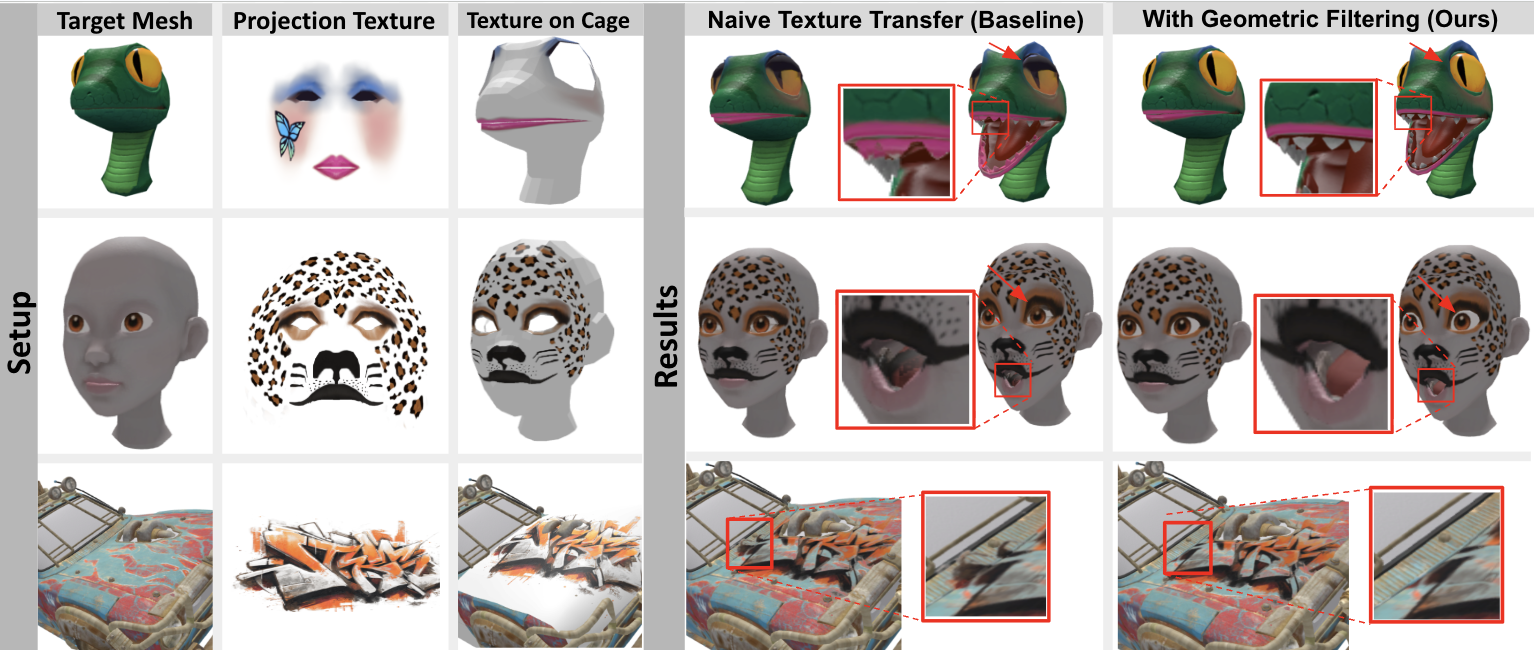}
  \caption{Cage-based texture transfer setups of a lizard, human, and car. Artifacts removed by the Geometric Filtering pipeline are highlighted in red.}
  \label{fig:teaserFramework}
\end{teaserfigure}

\maketitle

\section{Introduction}
Texture transfer, the process of projecting UV coordinates from a source mesh to a target mesh, is an essential process in real-time application suites. However, existing low-latency transfer methods \cite{lim2024projecting, 10.1145/1508044.1508091,lim2023reverse} frequently produce unacceptable artifacts because they project onto unintended vertices, which we define as Non-Cosmetic Zones (NCZs). These zones include internal geometry (e.g., a character's teeth and tongue) and specific segments the developer intends to exclude from projecting (e.g., the same character's eyes). Although prior research tackled identification of NCZs through photometric and edge guidance, these approaches are computationally intensive \cite{zhou2024ultravatar,10.1007/978-3-031-72980-5_11}, making them impractical for real-time mobile or interactive applications.
Alternative strategies for artifact suppression involve manual authoring by hand, a labor-intensive workflow. To bridge this gap, we present a real-time Geometric Filtering pipeline for artifact-free texture transfer. Our core innovation lies in the use of an auxiliary cage mesh, typically used for volumetric deformation \cite{feng2015avatar}, as a spatial reference. By leveraging the geometric relationship between the target mesh and the cage, we introduce a cage-guided filtering method that uses geometric ray casting and exposure ratios to cull NCZ artifacts. This Geometric Filtering approach enables the system to programmatically distinguish between valid surface areas and occluded or excluded zones without the overhead of heavy-weight models. To our knowledge, our Geometric Filtering framework represents a dual novelty: it is the first to repurpose cage-based spatial relationships for high-fidelity texture transfer; and it establishes a new benchmark for real-time transfer artifact suppression. By eliminating traditional projection artifacts with millisecond level latency, this framework significantly expands viability of real-time texture transfer for resource-constrained applications.



\section{Methodology}

Our Geometric Filtering pipeline begins by first specifying a cage mesh as the source for projecting UV mappings onto a target geometry. To resolve artifacts, we implement a filtering criterion based on spatial occupancy; specifically, we cast a ray from each target mesh vertex in the direction of the closest cage triangle's normal. These spatial queries, optimized via KD-Trees for both the target and cage meshes, provide the necessary data to identify NCZs. By leveraging these intersections as a geometric filter, our framework ensures a high-fidelity transfer. 


\begin{enumerate}[leftmargin=*]
    \item \textbf{Self-Intersections.} Internal target geometry is identified using self-intersection tests. For each target vertex $v$, we cast a ray in the direction of the normal $n$ of the nearest cage mesh triangle (\autoref{fig:targetIntersections}) from the vertex position $v_p$. If the ray $\vec{r}(t) = v_{p} + tn$ intersects the target mesh, this vertex is marked as an NCZ and we avoid transferring to it.

    \begin{figure}[h]
        \centering
        \begin{minipage}[b]{.45\textwidth}
          \centering
          \includegraphics[width=1.0\linewidth]{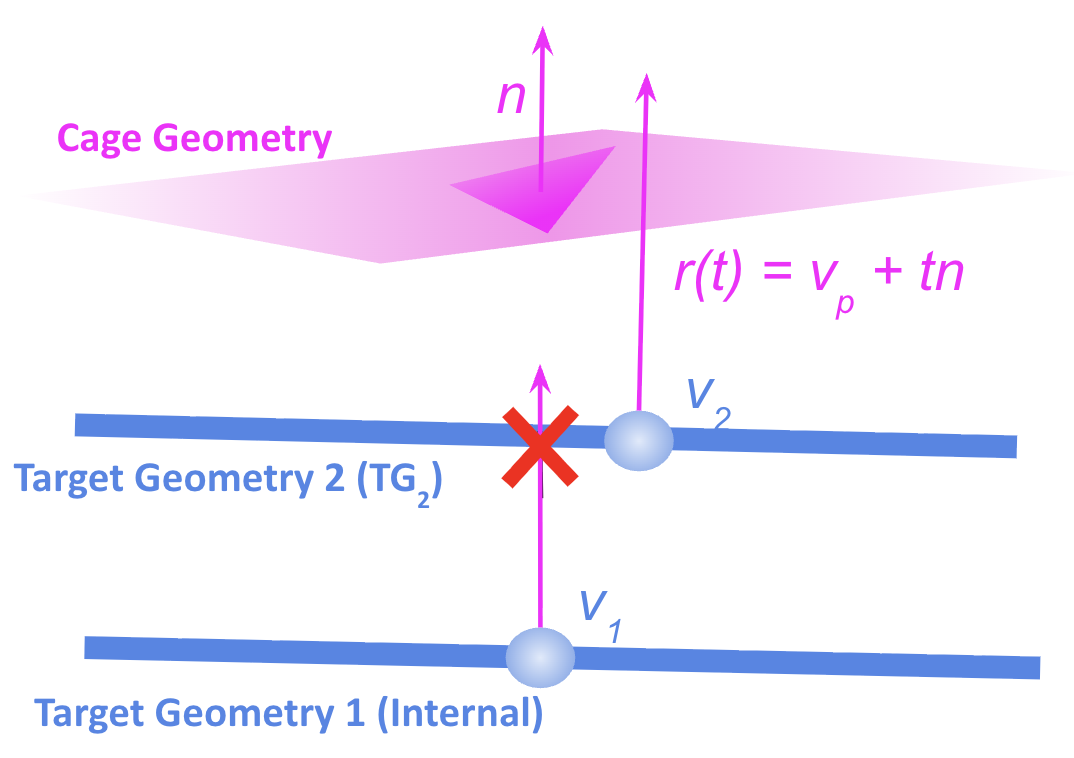}
          \captionof{figure}{Target vertex $v_1$ is classified as an NCZ because the ray projected from it intersects with target geometry ($TG_2$). Conversely, $v_2$ remains a valid candidate for transfer as it is clear of target mesh intersections.}
          \label{fig:targetIntersections}
        \end{minipage}
        \hfill
        \begin{minipage}[b]{.45\textwidth}
          \centering
          \includegraphics[width=1.0\linewidth]{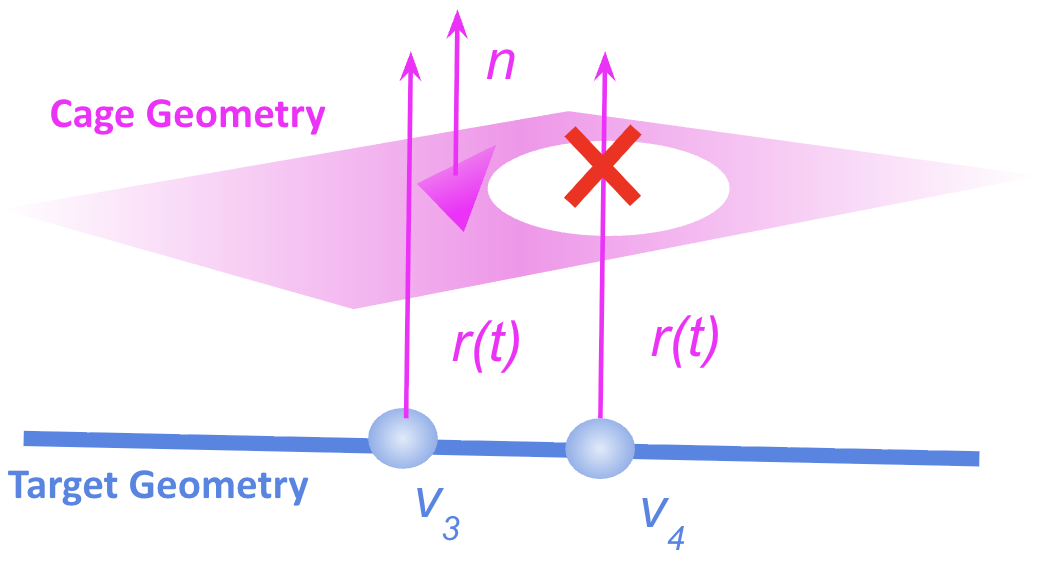}
          \captionof{figure}{Vertex $v_3$ intersects with the cage geometry, signifying a valid cosmetic area. Vertex $v_4$ misses the cage by passing through a hole, identifying it as an NCZ vertex.}
          \label{fig:cageIntersections}
        \end{minipage}
    \end{figure}
    
    
    \item \textbf{Cage Intersections.} We identify NCZ geometry specified by the developer through coverage of the cage. A cage-intersection check is performed for remaining vertices that still have UV transfer. By checking this subset, we avoid redundant calculations. Using the previously defined $\vec{r}(t)$, we test intersections with the cage. Vertices that \textit{do not} intersect the cage will be marked as NCZ (\autoref{fig:cageIntersections}).

    \item \textbf{Mesh Segmentation.} To facilitate the storage of segment-level metadata required for Geometric Filtering, the target mesh is partitioned into a set of discrete connected components $S=\{s_1, s_2, \dots, s_n\}$. Each component $s_i$ is defined as a maximal subgraph where any two vertices within $s_i$ are joined by at least one path, while remaining topologically disconnected from all other vertices in the mesh.
    \item \textbf{Threshold-Based Elimination} Rather than evaluating individual vertices in intersection tests, we analyze the collective transfer success of each mesh segment $s_i$. By calculating the fraction of a segment's total area that successfully received projection during the cage-intersection step, we can cull entire segments that fall below a threshold $F_s = C_s/E_s$. Here, $E_s$ represents the initial accumulated surface area of a segment after self-intersection filtering, while $C_s$ represents the remaining surface area after a subsequent cage-intersection pass. If a segment has a low $F_s$, we will mark all vertices in this segment as NCZ.


    
\end{enumerate}



\section{Results and Discussion}
We baseline our method against a Naive Texture Transfer approach, which reflects classical proximity-based mapping principles established in literature \cite{10.1145/1508044.1508091} that prioritize local spatial correspondence over intuitive results. Although this method achieves low latencies ($<50ms$), it is fundamentally limited by its inability to suppress artifacts in NCZs. Our visual results in \autoref{fig:teaserFramework} show that this naive approach leads to significant projection bleeding onto NCZs, which we have culled with our method. 
The gold standard approach for texture transfer is, of course, manual authoring. While artifact suppression is minimal with manual authoring, its scalability is low and latency is long.
\autoref{tab:ablation_comparison} compares our method against the Naive Transfer and Manual Authoring baselines.

\begin{table*}[h]
    \centering
    \caption{Comparison of Texture Transfer Methods across Fidelity, Performance, and Scalability of Figure \ref{fig:teaserFramework} Assets}
    \label{tab:ablation_comparison}
    \begin{tabular}{@{}lcccc@{}}
        \toprule
        \textbf{Method} & \textbf{Type} & \textbf{Latency} & \textbf{Artifact Suppression} & \textbf{Scalability} \\
        \midrule
        Manual Authoring & Artist-Driven & $\sim$60--120\,min & Gold Standard & Low \\
        Naive Transfer & Automated & $<$ 30--60\,ms & Low & High \\
        \textbf{Our Framework} & \textbf{Automated} & \textbf{$<$ 70--100\,ms} & \textbf{High} & \textbf{High} \\
        \bottomrule
    \end{tabular}
\end{table*}

Our geometric filtering framework achieves artifact suppression while maintaining a computational profile suitable for mobile deployment. Our methodology demonstrates efficiency across even low-end hardware tiers: on an Android Samsung Tablet S6 Lite, the process completes in 70ms for our 4,782-triangle lizard head (\autoref{fig:teaserFramework}). The runtime performance scales with the number of target vertices $V$, target triangles $N$, and cage triangles $M$. By utilizing KD-Trees to optimize spatial queries, we achieve an asymptotic complexity of $O(V \log(N + M))$. Memory consumption follows a linear scaling pattern $O(V + N + M)$; for instance, our test assets require a modest $\sim$20MB across all hardware tiers. This lean architectural footprint bypasses the resource-intensive scaling laws that govern modern neural approaches \cite{10.1007/978-3-031-72980-5_11}. While research into vision models demonstrates that performance gains are linked to massive memory consumption \cite{zhai2022scaling}, our framework maintains high-fidelity results with minimal memory costs.

\begin{figure}[h]
    \centering
    \begin{minipage}[b]{.5\textwidth}
      \centering
      \includegraphics[width=1.0\linewidth]{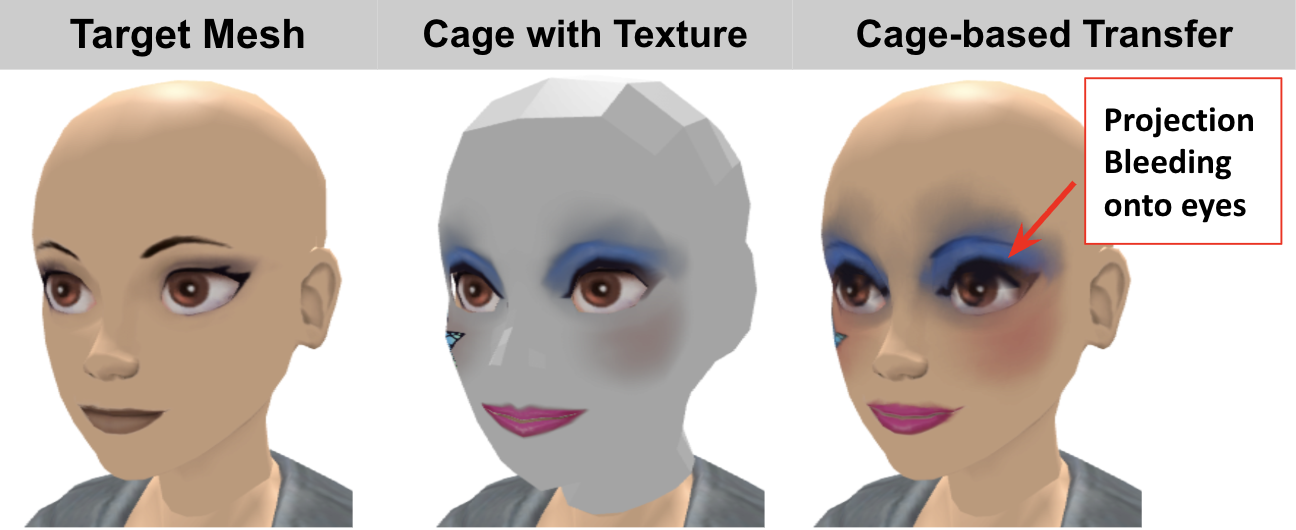}
      \captionof{figure}{The eyes are not disjoint sets of vertices from the rest of the target, letting minor black artifacts bleed in.}
      \label{fig:noSegmentation}
    \end{minipage}
    \hfill
    \begin{minipage}[b]{.5\textwidth}
      \centering
      \includegraphics[width=1.0\linewidth]{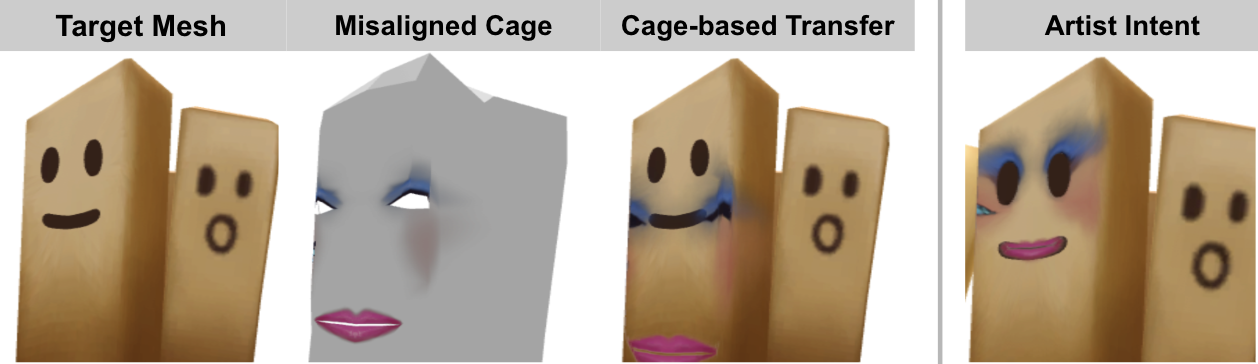}
      \captionof{figure}{This example demonstrates how a misaligned cage can neither filter geometry nor retain the artist intent.}
      \label{fig:badCage}
    \end{minipage}
\end{figure}

Limitations of cage-based transfer include a prerequisite for discretized mesh segmentation to identify NCZs, as well as sensitivity to the topological quality of the source cage. Without segmentation, the methodology relies exclusively on per-vertex intersection tests, which may not produce the exact results the developer intended (Figure \ref{fig:noSegmentation}). Although we plan to explore automated segmentation options, the efficacy of the framework remains fundamentally tied to the craftsmanship of the cage. Consequently, poorly fitted cages can lead to nonsensical results (Figure \ref{fig:badCage}). We imagine that AI models could be considered as an accessible solution to generate high quality cages.

To address the aforementioned drawbacks, traditional strategies (i.e., authoring secondary UV sets, hand-labeling NCZs) can serve as high-precision alternatives. Although creating a second UV set would yield perfect results, the workflow is unintuitive and labor-intensive \cite{chen2025artuvartiststyleuvunwrapping}. Similarly, manual NCZ labeling provides precision but introduces redundancy; since the transfer process already requires a source mesh, utilizing a cage as that source provides both the texture data and NCZ locality. Despite the dependency on cage craftsmanship and segmentation, our procedural pipeline bypasses the scalability bottlenecks in manual authoring.

By integrating rapid texture transfer with NCZ suppression, Geometric Filtering resolves a significant bottleneck in real-time asset interoperability. This technology expands the horizon for digital artistic expression, enabling a new class of highly customizable, artifact-free experiences in interactive media and game design.

\begin{acks}
Special thanks to Tomo Michigami, Rachel Yamada, Kelvin Lau, and Roblox's Avatar Personalization Team.
\end{acks}

\bibliographystyle{ACM-Reference-Format}
\bibliography{main}
\end{document}